\documentclass[9pt, a4paper]{article}
\usepackage{spconf,amsmath,graphicx}

\usepackage{xcolor}
\usepackage{mfirstuc}

\usepackage{comment}
\usepackage{amssymb}

\usepackage{times}
\usepackage{epsfig}
\usepackage{graphicx}
\usepackage{amsmath}
\usepackage{amssymb}
\usepackage{booktabs}
\usepackage{float}

\title{More Synergy, Less Redundancy: Exploiting Joint Mutual Information for Self-Supervised Learning}

\name{Salman Mohamadi, Gianfranco Doretto, Donald A. Adjeroh}
\address{Lane Department of Computer Science and Electrical Engineering, West Virginia University, USA}
\vspace{-10.em}

\begin{document}
%
\maketitle
%
\begin{abstract}
\vspace{-.7em}
Self-supervised learning (SSL) is now a serious competitor for supervised learning, even though it does not require data annotation.
Several baselines have attempted to make SSL models exploit information about data distribution, and less dependent on the augmentation effect. However, there is no clear consensus on whether maximizing or minimizing the mutual information between representations of augmentation views practically contribute to improvement or degradation in performance of SSL models. 
This paper is a fundamental work where, we investigate role of mutual information in SSL, and reformulate the problem of SSL in the context of a new perspective on mutual information.
To this end, we consider joint mutual information from the perspective of partial information decomposition (PID) as a key step in \textbf{reliable multivariate information measurement}. 
PID enables us to decompose joint mutual information into three important components, namely, unique information, redundant information and synergistic information.
Our framework aims for minimizing the redundant information between views and the desired target representation while maximizing the synergistic information at the same time. Our experiments lead to a re-calibration of two redundancy reduction baselines, and a proposal for a new SSL training protocol.
Extensive experimental results on multiple datasets and two downstream tasks show the effectiveness of this framework.
\end{abstract}
\vspace{-0.5cm}
\section{Introduction}
\vspace{-0.3cm}
Self-supervised learning (SSL) is among very successful principles that are needless of huge labeled datasets \cite{jing2020self}. While deep learning has shown tremendous success in many domains and applications including computer vision \cite{voulodimos2018deep}, biometrics \cite{mostofa2021deep}, genomics \cite{mohamadi2021human}, and etc, data-efficiency has been the focus of few problem domains such as deep active learning \cite{mohamadi2022deep,mohamadi2020deep}, and SSL \cite{chen2020simple}. Essentially, SSL frameworks consist of two key elements, namely, loss function, and pretext task \cite{mohamadi2023fussl}.
Basically, the pretext task is a proxy task which is to be solved using a supervisory signal from the unlabeled data, guided by an objective (loss) function \cite{mohamadi2023fussl}. Loss functions on the other hand generally guides learning the representation of a given sample by comparing two or multiple augmented views of the same sample with each other or with views of other samples. 
In fact, early baselines known as contrastive baselines were developed around the idea of contrasting augmented views of a sample with each other (positive pairs) and also with the views from other samples (negative pairs) \cite{oord2018representation,tian2020contrastive,he2020momentum,chen2020simple,bachman2019learning}. This type of baselines, however, suffer from the problem of potential representation collapse, as well as the need for large negative batches for effective 
representation. Next generation of baselines emerged as non-contrastive or negative pair-free baselines \cite{grill2020bootstrap,chen2021exploring}, essentially eliminating the need 
to contrast against negative views (negative pairs), and also almost with no risk of representation collapse. There is also a class of baselines known as clustering baselines such as \cite{caron2020unsupervised}, primarily based on clustering views of samples in the latent space. Two most recent baselines are based on redundancy reduction in representation of augmented views of the samples \cite{zbontar2021barlow,ermolov2021whitening}. This class of approaches mainly suggests that whitening the latent/embedding space of the a pair of networks trained on augmented views of samples allows for reducing redundant information in representation of the sample \cite{mohamadi2022deep}. Later theoretical work on whitening baselines showed that the prime reason for their success is eliminating another type of collapse, dimensional collapse \cite{hua2021feature,wen2022mechanism}.

In this work, {we assess how this whitening process unwittingly eliminates the synergistic information along with redundant information.} This relates to a larger  controversy on how mutual information relates to learning the target representation. Hence, in this paper, we start with investigating long-standing ambiguity about the role of mutual information in SSL. This eventually leads us to reconsider the problem of mutual information { between two variable (two views of a sample) }by reformulating it as joint mutual information {between three variables (two views and the target representation)}. To elaborate on the controversy, the general idea is to maximize the mutual information between encoder-representation of two augmented views for better representation; however some work \cite{rainforth2018tighter,tschannen2019mutual} suggested that more mutual information does not necessarily improve the representation. A recent work based on Info-Min principle suggests that, in fact, less mutual information between augmented views along with more task-associated information would improve the representation using a certain augmentation setting \cite{tian2020makes}. Another very recent work acknowledges the questionable role of mutual information, and suggests that decomposing the {estimation} of mutual information by adding an extra term representing the condition on the image with some blocked patches would reinforce the role of mutual information. However, this work is different from our work as they decompose the estimation of two-variable mutual information, whereas we 
focus on three-variable joint mutual information decomposition \cite{gutknecht2021bits}. In fact, we seek out the solution in the theory of partial information decomposition (PID). Eventually, this leads us to decompose the joint mutual information into its integral components, i.e., unique, redundant and synergistic component as was first introduced by \cite{williams2010nonnegative}. In the following, we first state the problem and discuss the decomposition of joint mutual information, then re-define SSL in this new context. We elaborate on the SSL baselines that rely on redundancy reduction, and propose a new training protocol for such SSL models, then empirically evaluate the new  protocol. 
\vspace{-0.81cm}
\section{Methods}
\vspace{-0.3cm}
\subsection{Problem Statement}
\vspace{-0.2cm}
From an information theoretic perspective, the general, though controversial, idea is that SSL frameworks generally tend to maximize the mutual information between encoder representation $f(.)$ of two augmented views $x_1$ and $x_2$ of sample data $x$ upper bounded by $I(x_1;x_2)$, i.e., $I(f(x_1);f(x_2))\leq I(x_1;x_2)$ \cite{sordoni2021decomposed, cover2006thomas}. This objective comes with challenges including how to optimally generate $x_1$ and $x_2$ \cite{tian2020makes} for actionable mutual information, as well as how to reduce redundant information in the representation \cite{ermolov2021whitening,zbontar2021barlow}. To elaborate on former challenge, Tian et al \cite{tian2020makes} suggested an heterodox idea, indicating that the augmentation process for generating views should be modified in a way that will enable reducing the mutual information between representation of positive views without affecting task-relevant information, i.e., mutual information is not necessarily task-relevant information. The later challenge, on the other hand, suggests that whitening the latent/embedding space would reduce redundant information. However, we argue that rather than focusing on mutual information between the representation of augmented views', the joint mutual information between views' representations and the target representation  could provide a possible way to resolve this controversy.
Hence, we take a totally different approach by formulating the core of SSL in terms of \textbf{joint mutual information \underline{ between views and the target representation }}.  
This leads us to the observation that, even though rigorous redundancy reduction through whitening such as in \cite{zbontar2021barlow} drops redundant information, it also risks reduction of useful synergistic information. This motivates us to design experiments to assess this claim in Sec. \ref{sec2.3}, and then to offer a training protocol to alleviate this loss of the synergistic element in joint mutual information. Specifically, we find it necessary to revisit the SSL principle from the joint mutual information perspective. Therefore, we assess two most recent baseline, Barlow-Twins \cite{zbontar2021barlow} and W-MSE \cite{ermolov2021whitening} which aim for redundancy reduction.
Below we elaborate on joint mutual information (in contrast with mutual information) and then we investigate two most recent baselines on whitening, which are also most relevant baseline to study redundancy and synergy.
\begin{figure}
\label{Fig1}
  \centering
  \includegraphics[scale=.18]{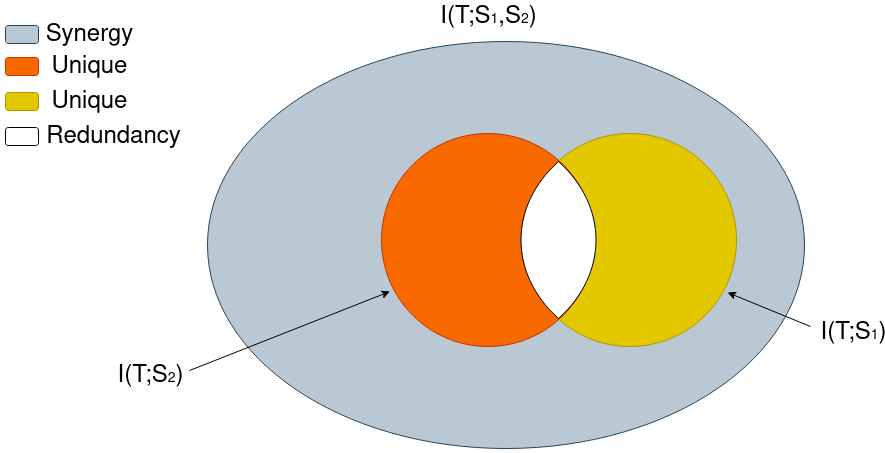}
\vspace{-1em}
		\caption{\scriptsize
		Partial information decomposition in case of three variables. 
\vspace{-3em}
}
\end{figure}
\vspace{-0.3cm}
\subsection{Decomposing Joint Mutual Information}
\vspace{-0.2cm}
For \textit{the first time ever} we consider the general SSL problem setting from the viewpoint of PID, which has diverse practical 
applications including in neuroscience, game theory and statistical learning. Hence, first we present the PID introduced in \cite{williams2010nonnegative} and then reformulate the SSL accordingly. We note that PID is not the only approach to multivariate measurement of information. However, it has multiple advantages in our SSL context, including non-negative decomposition of information as well as separate and simultaneous measurement of redundancy and synergy as distinct quantities \cite{timme2014synergy}. 
This new interpretation of SSL is primarily posed to address the ambiguity in the role of mutual information in SSL. 

The PID is an approach to a non-overlapping decomposition of the joint mutual information between two sets of variables, a set of two or more source variables carrying information about a target, as well as the single target variable. This decomposition has been challenging as the proposed solutions mostly consisted of negative information terms, until a breakthrough work by \cite{williams2010nonnegative} which introduced a non-negative decomposition in terms of quantifying three components, the  unique, redundant, and synergistic information.

In its simplest form, suppose we have two source variables $S_1$ and $S_2$ carrying joint mutual information $I(T; S_1,S_2)$ about a target variable $T$. Hence each of the source variables has mutual information with the target variable. Decomposing the joint mutual information into some non-negative components, models information interaction to assess the contribution offered by each source variable and combination of sources. According to \cite{williams2010nonnegative}, as shown in Fig. \ref{Fig1} the joint mutual information between sources and target, could be decomposed as three elements, unique, redundant, and synergistic information. Unique information is the part provided by each source separately, redundant information is the minimum information provided by each source (aka common mutual information), and synergistic information is the information provided only by a combination of $S_1$ and $S_2$ about $T$, which neither alone can provide \cite{gutknecht2021bits}.
\begin{equation}
    \label{eq.PID}
    \scriptsize
    \begin{split}
        & I(S_1,S_2:T)=\text{Redundancy}(T; S_1, S_2) + \\
        & \text{Synergy}(T; S_1, S_2)+ \text{Unique}(T; S_1) + \text{Unique}(T; S_2)
    \end{split}
\end{equation}
Now consider the general setting of SSL, where at least two random augmented views of a sample are generated. The goal is to contrast them in order to learn a representation that is maximally informative about the original sample distribution, while minimally 
informative about the augmentation. This contrast in essence {creates an information interaction} between the information of the variables which could be studied under the PID framework. Here, the two augmented views could be seen as source variables $S_1$ and $S_2$, whereas the original sample distribution is the target variable $T$. In a more general sense, $T$ could be considered the class distribution representing the invariant representation of the views of a given sample, i.e., the class the data sample belongs to. 
Here,  as only redundant and synergistic information will be the results of interaction in contrasting views in SSL frameworks, unique information is not the subject matter of our study in this work. Unique information would be the subject of  non-contrastive supervised learning on labeled data.
\vspace{-0.3cm}
\subsection{Redundancy Reduction Baselines}
\vspace{-0.2cm}
Interestingly, two most recent SSL baselines \cite{ermolov2021whitening,zbontar2021barlow} are redundancy reduction (aka hard/soft whitening) baselines. Both baselines take advantage of whitening (Cholskey whitening) of latent/embedding space of a cross-correlation matrix computed from augmented views of the same sample. Ermolov et al \cite{ermolov2021whitening} proposed a hard whitening method based on a recent version of Cholesky decomposition \cite{dereniowski2003cholesky, siarohin2018whitening} for whitening the latent space vectors. At the same time, Zbontar et al  \cite{zbontar2021barlow} has gained more popularity by proposing as simpler process called soft whitening, which essentially forces the cross-correlation matrix of the embedding vectors of two networks to identity matrix. The later approach, known as Barlow-Twins, suggests that their whitening approach intuitively results in redundancy reduction embedded in off-diagonal elements of the cross-correlation matrix.
We use both approaches for our investigation, and provide further insight on the synergy versus redundancy. However due to the lack of space we only represent the theoretical reformulation of Barlow-Twins under our framework, as it is more popular. The following is the loss function of Barlow-Twins:
\begin{equation}
    \label{eq6}
    \scriptsize
    \quad \mathcal{L}_{BT} \triangleq \sum_{i}(1-C_{ii})^2 + \lambda\sum_{i}\sum_{j\neq i}(C_{ij})^2
\end{equation}
\begin{equation}
    \label{eq7}
    \scriptsize
 C_{ij}\triangleq \frac{\sum_m z_{m,i}^{A} z_{m,j}^{B}}{\sqrt{\sum_m (z_{m,i}^{A})^2}\sqrt{\sum_m (z_{m,j}^{B})^2}}
\end{equation}
where $C_{ij}$ are elements of the cross-correlation matrix $C$ between the embedding vectors with element $z$ of two networks (twins), as presented in Eq. \ref{eq7}. $\lambda$ as a weighting factor, originally set to $5\times 10^{-3}$.
\vspace{-0.5cm}
\subsection{Assessing synergy and redundancy}
\vspace{-0.2cm}
\label{sec2.3}
In order to lay a context for PID in the SSL context, we find it necessary to design simple experiments around redundancy reduction and synergy in Barlow-Twins (BT). Note that as the augmented views for a sample generated under standard augmentation for SSL share lots of information in common (redundant or commonly known as mutual information), BT attains desirable performance by implementing rigorous redundancy reduction. However we argue that if the redundant information was not as much, the performance would drop sharply. To assess this, we apply  heavy augmentation on samples (such as \cite{bai2022directional}) to generate views with 
significantly less redundant information, and then test BT performance on these. The top-1 accuracy for CIFAR10 and CIFAR100 (under experimental settings in next section) drops by $\%5.69$ and $\% 5.13$ respectively. Now under same heavy augmentation, we re-calibrate BT by setting $\lambda=0.1$ and also forcing off-diagonal elements to a multivariate Gaussian $\mathcal{N}(0,1)$ rather than zero to allow them better affect the learned representation, we gain accuracy, $\%0.91$, and $\%0.81$ compared with the former case. This  implies that the off-diagonal elements not only carry redundant information, but also some other type of information. Otherwise allowing more redundancy by using multivariate Gaussian off-diagonal elements would have degraded the performance.  We argue that off-diagonal elements do not only represent redundant information, \textbf{but also synergistic information}. This is why when we reduce the redundant information by implementing heavy augmentation, BT's rigorous redundancy reduction constraint on off-diagonal elements of the cross-correlation matrix, degrades the performance by targeting synergistic information. Below, we propose a training protocol that works even better than forcing off-diagonal elements to multivariate Gaussian, and present our experimental results on two baselines BT and W-MSE in Sec. \ref{exp} to show the generality of our framework.
\vspace{-0.45cm}
\section{Synergy-based training protocol}
\vspace{-0.2cm}
We aim for re-calibrating the redundancy reduction in BT \cite{zbontar2021barlow} and W-MSE \cite{ermolov2021whitening} toward protecting the most synergistic information during the redundancy reduction process. In its current form, BT approach does not seem to optimally reduce redundancy, without significant loss in the synergistic component. 
Our approach {consists of a serial pre-training with first phase of dropping redundancy and second phase of adding to synergy.} Hence, in this section, we define a new training protocol aiming for extracting more synergistic information during the process of redundancy reduction which will be implemented on both BT and W-MSE. We present this protocol 
aimed at more synergy and less redundancy via the use of engineered off-diagonal elements, to show  the effectiveness of the joint mutual information decomposition in SSL. As the augmented views of a sample under standard augmentation share lots of mutual information, we find it practically more efficient to update/replace the loss function of BT and W-MSE after initial pre-training with the original loss function which solely aims at redundancy reduction. This is done under a new training protocol with two phases of pre-training in two different settings. First phase aims at reducing the redundancy, while the second phase aims at adding to synergy. Below we only present the new formulation for BT, however, we provide the experimental results for both BT and W-MSe.\\
\indent \textbf{A. Gaussian off-diagonal:} After initial pre-training of original model, here BT, the network is fixed, to resume the training with an updated loss. For BT we set $\lambda=0.1$ and replace the second term in Eq. \ref{eq6} with $\lambda\sum_{i}\sum_{j\neq i}(C_{ij}-G_{ij})^2$ where $G_{ij}$ are the multivariate Gaussian elements of a square matrix $G$ of proper size. This allows the BT to better consider the off-diagonal elements of the cross-correlation matrix, which convey synergy and redundancy.\\
\indent \textbf{B. Reinforced off-diagonal:} After initial pre-training of original model, here BT, the network is fixed and the average $C^{Ave}_{ij}=\frac{1}{n}\sum_n C_{ij}$ over all $n$ samples will be computed. Then training resumes with new $\lambda=0.1$ and the second term in Eq. \ref{eq6} updated as $\lambda\sum_{i}\sum_{j\neq i}(C_{ij}-C^{Ave}_{ij})^2$ forcing each off-diagonal element to its corresponding average.
\vspace{-0.4cm}
\section{Experiments and Results}
\vspace{-0.3cm}
\label{exp}
\subsection{Experiments}
\vspace{-0.21cm}
\textbf{Baselines:} Our modification on BT and W-MSE 
\cite{zbontar2021barlow,ermolov2021whitening} resulted in GSBT and RSBT, as well as GSW-MSE and RSW-MSE respectively. We perform experiments using our new training protocol under standard and heavy data augmentation. We contrast it with most recent baselines including Whitening-MSE ($d=4$) \cite{ermolov2021whitening}, a non-contrastive baseline BYOL \cite{grill2020bootstrap}, and a clustering-based baseline SwAV \cite{caron2020unsupervised}. Following \cite{ermolov2021whitening}, latent spaces of all methods are 
$L_2$-normalized.\\
\indent \textbf{Dataset and augmentation:} We use six datasets including ImageNet \cite{deng2009imagenet}, CIFAR10, CIFAR100 \cite{krizhevsky2009learning}, Tiny ImageNet \cite{le2015tiny}, ImageNet-100, and VOC0712. We use two sets of augmentation protocols, standard and heavy. For standard augmentation including random grayscaling, random crop, color jittering, aspect ratio adjustment, and horizontal mirroring,  we follow the settings in \cite{chen2020simple}, and for heavy augmentation we follow the settings in \cite{bai2022directional}.\\
\indent \textbf{Network \& implementation details:} For CIFAR10/100, following the details of each baseline \cite{chen2020simple,grill2020bootstrap,chen2021exploring,caron2020unsupervised,ermolov2021whitening,zbontar2021barlow}, we use ResNet18 while for ImageNet, Tiny ImageNet, and VOC0712 we use ResNet50 \cite{he2016deep} , for the encoder and the same projector head as \cite{zbontar2021barlow}, with the same size of projector output in all baselines. For VOC0712 similar to \cite{zbontar2021barlow}, Faster R-CNN \cite{ren2015faster} is used. Optimization of all experiments were done using Adam optimizer \cite{kingma2014adam}. Pre-training of RSBT, GSBT as well as RSW-MSE and GSW-MSE are performed in two phases, a phase one (redundancy reduction) consists of 500 epochs with batch size of 1024, which starts with a learning rate of $0.15$ for some 20 epochs and drops to $0.001$ for the remaining epochs. Phase two (synergy addition) also consists of another 500 epochs with the learning rate of $0.001$, with their modified loss functions. The weight decay in both phases and all other experiments is $10^{-6}$.
\vspace{-0.5cm}
\subsection{Evaluation and results}
\vspace{-0.4em}
Similar to former methods, we perform the standard supervised linear evaluation for classification task as well as detection. Classification involves fixing the encoder weights after pre-training and replacing the projector with a linear classifier (fully connected followed by softmax), and training the linear classifier for some 500 epochs on evaluation data, and then testing it. The classification resluts for ImageNet, CIFAR10/100, Tiny ImageNet, and ImageNet-100 with different settings of proposed training protocol are presented in the Tables 1, 2, and 3, whereas the detection results with VOC0712 is presented in Table 1. Results for modified BT using our protocol is presented in Table 1 and 2, whereas the results for modified W-MSE using our protocol is available in Table 3. In both settings of data augmentation, our method outperforms prior approaches. While heavy augmentation degrade the performance of other approaches, it even improves the RSBT, GSBT, as well as RSW-MSE and GSW-MSE which shows  robustness of our approach.
\begin{table}[H]
\vspace{-.5cm}
  \label{table1}
  \caption{
  \footnotesize
  Results of our methods on 2 downstream tasks -- classification and object detection.  Top-1  classification accuracy with supervised linear evaluation on ImageNet (Left), and Tiny ImageNet (Middle). Object detection results for VOC0712 (Right). Used both standard and heavy augmentations. 
  }
  \centering
  \scriptsize
  \begin{tabular}{p{1.4cm} p{.7cm}|p{1.cm}||p{.7cm}|p{1.1cm}||p{.6cm}|p{1.1cm}}
    \toprule
    Framework   & \multicolumn{2}{c}{ ImageNet (Class.)} & \multicolumn{2}{c}{Tiny ImageNet (Class.)}  & \multicolumn{2}{c}{VOC0712 (Det.)}    \\
    \cmidrule(r){2-7}
      &  Standard   & Heavy  & Standard   & Heavy  & A$_{All}$ & A$_{50}$\\
    \cmidrule(r){1-7}
      BYOL  & \textbf{74.3}  & 60.5 \: ($\downarrow $ ) & 51.43  & 47.16 \: ($\downarrow $ ) & 56.8 & 82.5 \\
       SwAV  & 71.8 & 58.9 \: ($\downarrow $ ) & 51.03  & 44.25 \: ($\downarrow $ ) &56.1 & \textbf{82.6}\\
    W-MSE4  & 73.1 & 61.6 \: ($\downarrow $ )&  50.59 & 48.11 \: ($\downarrow $ )  & \textbf{56.9} & 82.4\\
   {B-Twins}  &  {73.2} &61.9 \: ($\downarrow $ )&  {50.63} &47.49 \: ($\downarrow $ ) & 56.8 & \textbf{82.6} \\
   \midrule
   \textbf{GSBT }  &  \textbf{75.4} &\textbf{75.5} \: ($\uparrow $ ) &   \textbf{51.54} &\textbf{52.08} \: ($\uparrow $ )  &\textbf{57.3} & \textbf{82.6}\\
   \textbf{RSBT }  &\textbf{76.1} &\textbf{76.5} \: ($\uparrow $ )  &   \textbf{51.94} &\textbf{52.46} \: ($\uparrow $ ) &\textbf{57.8} & \textbf{82.7} \\
    \bottomrule
  \end{tabular}
\end{table}
\vspace{-0.81cm}
\begin{table}[H]
  \label{table2}
  \caption{
  \footnotesize
  Top-1 classification accuracy with supervised linear evaluation for CIFAR10/100  under both standard and heavy augmentations.} 
  \centering
  \scriptsize
  \begin{tabular}{p{1.4cm} p{1.5cm} p{1.5cm}| p{1.5cm} p{1.5cm}}
    \toprule
    Framework   & \multicolumn{2}{c}{CIFAR10} & \multicolumn{2}{c}{CIFAR100}  \\
    \cmidrule(r){2-5}
        & Standard   & Heavy & Standard & Heavy \\
    \cmidrule(r){1-5}
      BYOL  &  91.81  & 84.11\: ($\downarrow $ )&    66.65 &59.93\: ($\downarrow $ ) \\
       SwAV  &  92.11  & 85.25 \: ($\downarrow $ )& 67.77  & 60.89\: ($\downarrow $ )\\
    W-MSE4  &  91.89 & 87.74\: ($\downarrow $ )&  67.58  &62.14 \: ($\downarrow $ )\\
   {B-Twins}  &   {92.33} &{86.64} \: ($\downarrow $ ) & {67.47}  &62.34\: ($\downarrow $ ) \\
   \midrule
    \textbf{GSBT }  &   \textbf{92.83} &\textbf{93.10}\: ($\uparrow $ ) & \textbf{67.81} & \textbf{68.23}\: ($\uparrow $ )\\
     \textbf{RSBT }  &   \textbf{93.03} &\textbf{93.47}\: ($\uparrow $ ) & \textbf{68.11} & \textbf{68.83}\: ($\uparrow $ )\\
    \bottomrule
  \end{tabular}
  \vspace{-0.5cm}
\end{table}
\begin{table}[H]
\vspace{-.75em}
  \label{table1}
  \caption{
  \footnotesize
  Experiments on another baseline, W-MSE, Top-1  classification accuracy  with supervised linear evaluation for CIFAR100 and ImageNet-100.}
  \centering
  \scriptsize
  \begin{tabular}{p{1.2cm} p{1.2cm}|p{1.2cm}||p{1.2cm}|p{1.2cm}}
    \toprule
    Framework   & \multicolumn{2}{c}{CIFAR100} & \multicolumn{2}{c}{ImageNet100}  \\
    \cmidrule(r){2-5}
        & Standard   & Heavy & Standard   & Heavy           \\
    \cmidrule(r){1-5}
    W-MSE4  &   67.58 & 62.14 \: ($\downarrow $ ) & \textbf{79.02} & 71.14 \: ($\downarrow $ )   \\
   {B-Twins}  &   67.47 &62.34 \: ($\downarrow $ ) & 77.93 & 72.57 \: ($\downarrow $ ) \\
   \midrule
   \textbf{GSW-MSE}  &   \textbf{68.26} &\textbf{68.51} \: ($\uparrow $ ) &  \textbf{79.58} & \textbf{79.91} \: ($\uparrow $ ) \\
   \textbf{RSW-MSE}  &   \textbf{69.11} &\textbf{69.32} \: ($\uparrow $ ) & \textbf{79.93} &\textbf{80.66} \: ($\uparrow $ )\\
    \bottomrule
  \end{tabular}
\end{table}
\vspace{-0.45cm}
\section{Conclusion}
\vspace{-0.3cm}
We address the ambiguity regarding how mutual information relates to better representation in SSL. To this end, we explore the use of PID in SSL and we re-define the formulation of SSL problem in terms of joint mutual information between three variables (two views of a sample and its original representation). This allows for recognition of synergistic information along with the redundant information and their role in boosting performance. We design and perform extensive experiments on the most recent redundancy reduction baselines, BT and W-MSE and instantiate the theoretical solution in practice under a new training protocol. 
\bibliographystyle{IEEEbib}
\scriptsize
\bibliography{strings,refs}

\end{document}